\documentclass[10pt,twocolumn,letterpaper]{article}

\usepackage{iccv}
\usepackage{times}
\usepackage{epsfig}
\usepackage{graphicx}
\usepackage{amsmath}
\usepackage{amssymb}
\usepackage{array}
\usepackage{float}
\usepackage{subfigure}

\usepackage[breaklinks=true,bookmarks=false]{hyperref}

\iccvfinalcopy 

\def\iccvPaperID{7496} 

\ificcvfinal\fi

\begin{document}

\title{BCSSN: Bi-direction Compact Spatial Separable Network for Collision Avoidance in Autonomous Driving}

\author{Haichuan Li \and Liguo Zhou \and Alois Knoll\\
\and
Chair of Robotics, Artificial Intelligence and Real-Time Systems\\
Technical University of Munich\\
{\tt\small haichuan.li@tum.de, liguo.zhou@tum.de, knoll@in.tum.de}
}

\maketitle
\ificcvfinal\thispagestyle{empty}\fi

\begin{abstract}
Autonomous driving has been an active area of research and development, with various strategies being explored for decision-making in autonomous vehicles. Rule-based systems, decision trees, Markov decision processes, and Bayesian networks have been some of the popular methods used to tackle the complexities of traffic conditions and avoid collisions. However, with the emergence of deep learning, many researchers have turned towards CNN-based methods to improve the performance of collision avoidance. Despite the promising results achieved by some CNN-based methods, the failure to establish correlations between sequential images often leads to more collisions. In this paper, we propose a CNN-based method that overcomes the limitation by establishing feature correlations between regions in sequential images using variants of attention. Our method combines the advantages of CNN in capturing regional features with a bi-directional LSTM to enhance the relationship between different local areas. Additionally, we use an encoder to improve computational efficiency. Our method takes "Bird's Eye View" graphs generated from camera and LiDAR sensors as input, simulates the position (x, y) and head offset angle (Yaw) to generate future trajectories. Experiment results demonstrate that our proposed method outperforms existing vision-based strategies, achieving an average of only 3.7 collisions per 1000 miles of driving distance on the L5kit test set. This significantly improves the success rate of collision avoidance and provides a promising solution for autonomous driving.
\end{abstract}

\section{Introduction}

Trajectory prediction~\cite{wang2018trajectory,liu2018hybrid,rodrigues2017multiagent} is a critical task in autonomous driving for collision avoidance, as it requires \begin{figure}[ht]
    \centering
    \subfigure[Front Collision]{
    \includegraphics[width=0.48\textwidth]{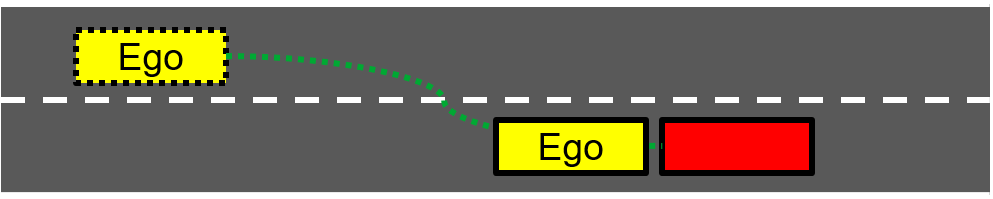}
    }
    \subfigure[Side Collision]{
    \includegraphics[width=0.48\textwidth]{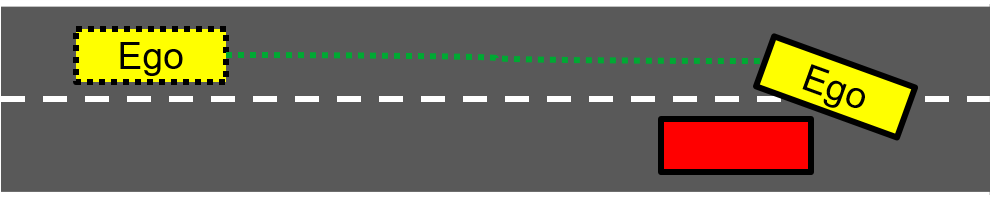}
    }
    \subfigure[Rear Collision]{
    \includegraphics[width=0.48\textwidth]{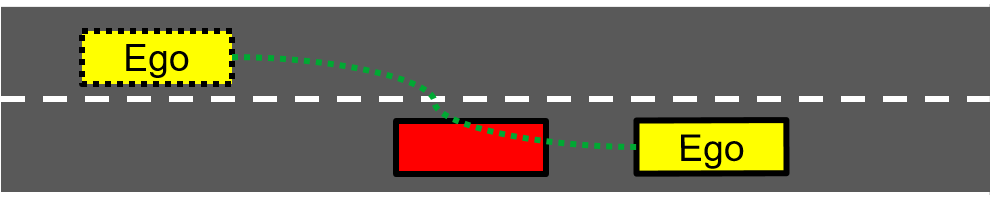}
    }
    \caption{Different collision cases during driving}
    \label{fig:collision}
\end{figure}

\begin{figure*}[ht]
    \centering
    \includegraphics[width=\textwidth]{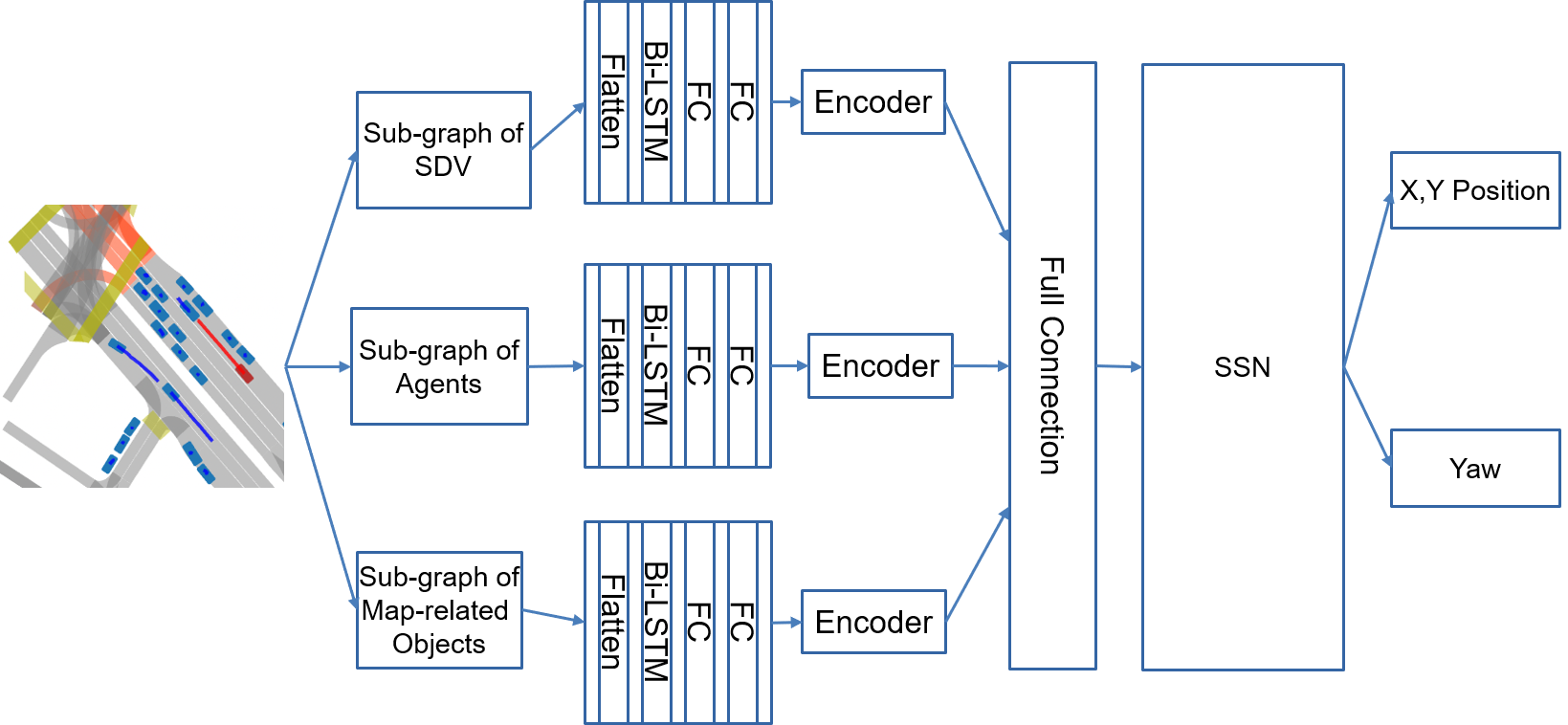}
    \caption{Overview of Bi-direction Compact Spatial Separable Network }
    \label{fig:CSSN}
\end{figure*} 
\noindent the autonomous vehicle (AV) to accurately estimate the future motion of surrounding objects and plan its own motion accordingly shows in Fig~\ref{fig:collision}. This task is challenging due to the uncertainty of the surrounding environment and the complexity of the motion patterns of other vehicles, pedestrians, and cyclists.
To improve the accuracy of trajectory prediction, various techniques have been proposed. For instance, multi-modal prediction~\cite{deo2020end, zhao2019multi} and uncertainty estimation~\cite{gal2016dropout, bhattacharyya2020latent} have been utilized to consider multiple possible future trajectories and their probabilities. Attention mechanisms~\cite{kim2019deterministic, ma2019trafficpredict} and graph-based models~\cite{zhao2019multi, yu2021grnet} have been proposed to capture the importance and relationships among different objects.
Convolutional Neural Network (CNN) has made outstanding contributions to vision tasks and has been widely applied to traffic scenes due to its excellent regional feature extraction capabilities. Based on this advantage, CNN will obtain the local feature information of sequential related frames in our network. Additionally, since the motion trajectory is planned for AV, which means each position point has a sequential relation (the later points depend on former points). It is necessary to establish the relationship between each local feature of the image obtained by CNN. To address this, some strategies use CNN plus RNN to deal with sequential graphs as input, such as STDN~\cite{yao2018modeling}, CRNN~\cite{shin2017crnn}, LSTM-CNN~\cite{ballas2015lstm_cnn}, RCNN~\cite{chen2017rcnn}.

Although the above strategies have performed well in a large number of vision tasks, their performances are still far inferior to similar-sized convolutional neural networks counterparts, such as EfficientNets~\cite{tan2019efficientnet} and RepVGG~\cite{ding2021repvgg} in Fig.\ref{othernetwork}. We believe this is due to the following aspects. First, the huge differences between the sequential tasks of NLP and the image tasks of CV are ignored. For example, when the local feature information acquired in a two-dimensional image is compressed into one-dimensional time series information, how to achieve accurate mapping becomes a difficult problem. Second, it is difficult to keep the original information of inputs since after RNN layers, we need to recover the dimension from one to three. Besides, due to the several transformations between different dimensions, that process becomes even harder, especially our input size is 224×224×5. Third, the computational and memory requirement of switching between layers are extremely heavy tasks, which also becomes a tricky point for the algorithm to run. Higher hardware requirements as well as more running time arise when running the attention part.

In this paper, we propose a new network structure based on CNN, Bi-LSTM, encoder, and attention to trajectory-generating tasks in autonomous driving. The new network structure overcomes these problems by using Bi-direction Compress Sequential Spatial Network (BCSSN). As shown in Fig.~\ref{fig:CSSN}, input Bird's Eye View (BEV) images first will be divided into three sub-graphs. Then go through bi-directional frame-related(BIFR) blocks which consist of flatten layer, Bi-LSTM layer, and full connect layers. After that, information will be fed into the main stem, the convolution stem for fine-grained feature extraction, and are then fed into a stack of SSN (Sequential Spatial Network) blocks in Fig.~\ref{fig:SSN} for further processing. The Upsampling Convolutional Decreasing (UCD) blocks are introduced for the purpose of local information enhancement by deep convolution, and in SSN block of features generated in the first stage can be less loss of image resolution, which is crucial for the subsequent trajectory adjustment task. 

In addition, we adopt a staged architecture design using three convolutional layers with different kernel sizes and steps gradually decreasing the resolution (sequence length) and flexibly increasing the dimensionality. Such a design helps to extract local features of different scales and, since the first stage retains high resolution, our design can effectively reduce the resolution of the output information in the first layer at each convolutional layer, thus reducing the computational effort of subsequent layers. The Reinforcement Region Unit (RRU) and the Fast MultiHead Self-Attention (FMHSA) in the SSN block can help obtain global and local structural information within the intermediate features and improve the normalization capability of the network. Finally, average pooling is used to obtain better trajectory tuning. 

Extensive experiments on the lykit dataset~\cite{pmlr-v155-houston21a} demonstrate the superiority of our BCSSN network in terms of accuracy. In addition to image classification, SSN block can be easily transferred to other vision tasks and serve as a versatile backbone.

\section{Related Works}

Rule-based systems\cite{harper1993rule,yilmaz2004rules,hinkelmann2004rule}:
Rule-based systems are a type of artificial intelligence that uses a set of rules to make decisions. These rules are typically expressed in if-then statements and are used to guide the system's behavior. Rule-based systems have been used in a variety of applications, including expert systems, decision support systems, and automated planning systems.

Decision trees\cite{breiman1984classification,mehta1996using,veropoulos1999controlling}:
Decision trees are a type of machine learning algorithm that is used for classification and regression analysis. They are built using a tree structure, where each internal node represents a decision based on a feature, and each leaf node represents a prediction. Decision trees are widely used in various fields, including business, medicine, and engineering. 

Markov Decision Processes\cite{puterman2014markov,mnih2016asynchronous,ross2014reinforcement,lemon2019multi} (MDPs):
Markov Decision Processes are a mathematical framework for modeling sequential decision-making problems, where an agent interacts with an environment in discrete time steps. The framework involves a set of states, actions, rewards, and transition probabilities that govern the agent's behavior. MDPs have been used in a wide range of applications, such as robotics, finance, healthcare, and transportation.

Bayesian networks\cite{koller2009probabilistic,jensen2001bayesian,heckerman1995learning}: A Bayesian network is a probabilistic graphical model that represents a set of random variables and their conditional dependencies using a directed acyclic graph. It is a powerful tool for probabilistic inference, learning from data, and decision making under uncertainty.

\begin{figure}[!ht]
    \centering
    \subfigure[ResNet-50]{
    \includegraphics[height=3.5cm]{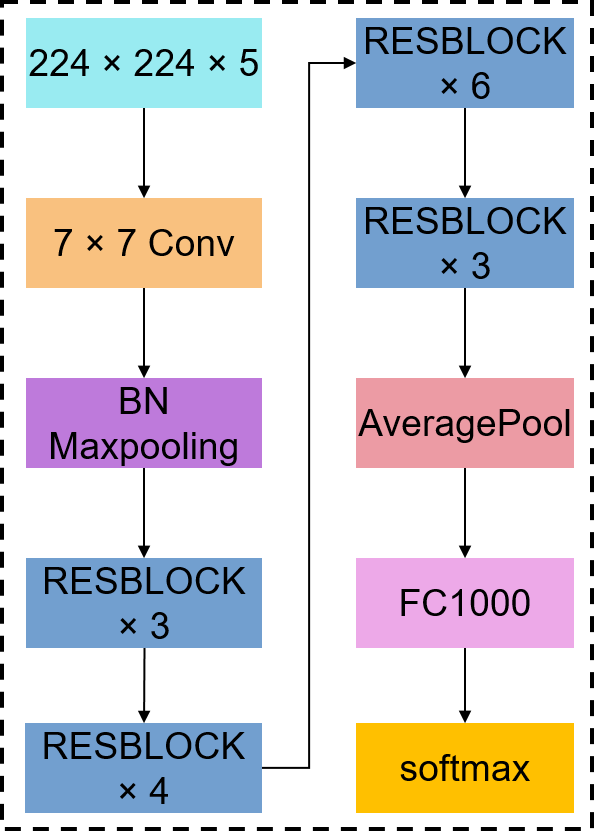}
    }
    \subfigure[RepVGG]{
    \includegraphics[height=3.5cm]{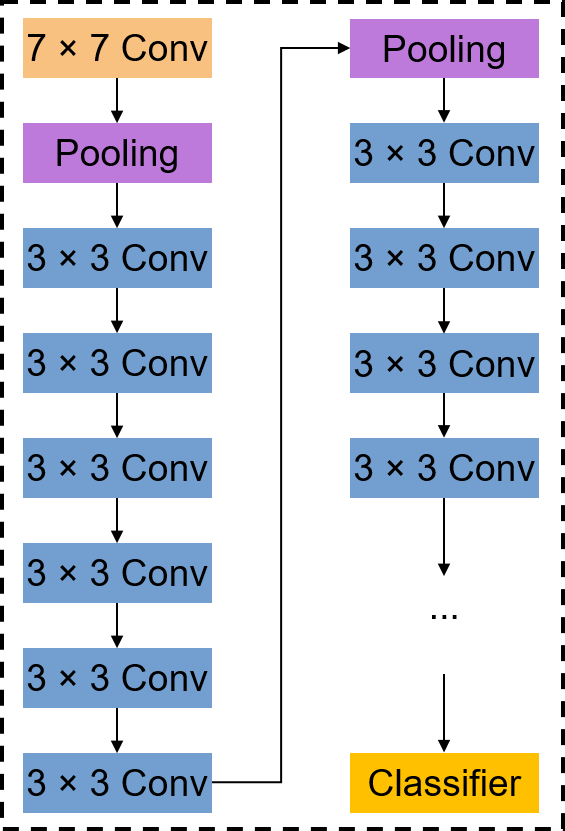}
    }
    \subfigure[ViT]{
    \includegraphics[height=3.5cm]{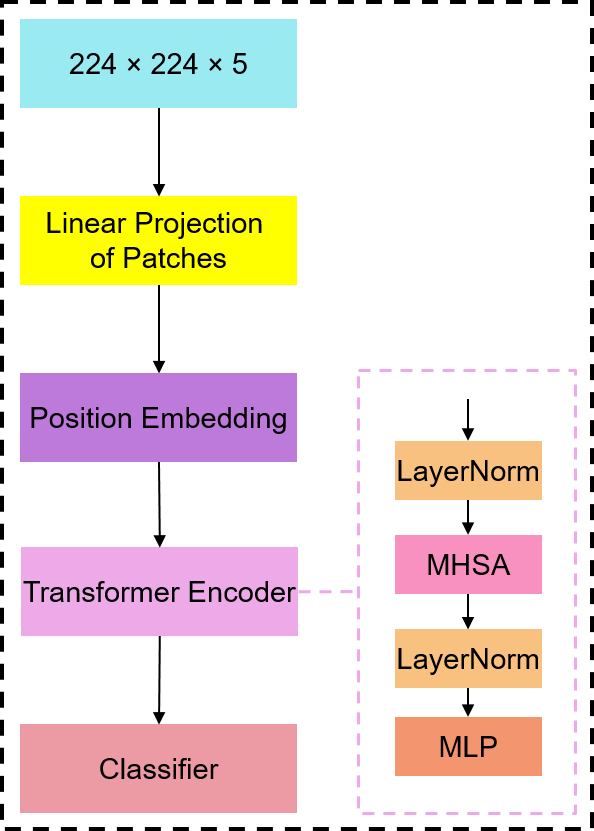}
    }
    \caption{Three popular network structures in vision areas. The structure of ResNet-50 is shown in (a). The structure of RepVGG is shown in (b). The structure of ViT is shown in (c).}
    \label{othernetwork}
\end{figure}

Except for the mentioned four methods, there are some popular strategies as well. Over the past decade, autonomous driving has flourished in the wave of deep learning, where a large number of solution strategies are based on computer vision algorithms, using images as the primary input. The prevailing visual neural networks are typically built on top of a basic block in which a series of convolutional layers are stacked sequentially to capture local information in intermediate features. However, the limited receptive field of the small convolution kernel makes it difficult to obtain global information, which hinders the high performance of the network on highly feature-dependent tasks (such as trajectory prediction and planning). In view of this dilemma, many researchers have begun to deeply study self-attention-based~\cite{vaswani2017attention} networks with the ability to capture long-distance information. Here, we briefly review traditional CNN and recently proposed visual networks. Convolutional neural network. The first standard CNN was proposed by LeCun~\cite{lecun1989backpropagation} et al. and was used for handwritten character recognition. Based on this foundation, a large number of visual models have achieved cross-generational success in a variety of tasks with images as the main input. Google Inception Net ~\cite{szegedy2015going} and DenseNet ~\cite{huang2017densely} showed that deep neural networks consisting of convolutional and pooling layers can yield adequate results in recognition.  ResNet~\cite{he2016deep} in Fig.\ref{othernetwork} is a classic structure that has a better generalization ability by adding shortcut connections to the underlying network. To alleviate the limited acceptance domain in previous studies, some studies used the attention mechanism as an operator for adapting patterns.

Besides, several novel visual networks have been proposed recently, which have achieved remarkable performance in various computer vision tasks. For example, the Mask R-CNN~\cite{he2017mask} proposed by He et al. extends the Faster R-CNN framework with an additional mask branch to perform instance segmentation. SENet~\cite{hu2018squeeze} and MobileNetV3 ~\cite{howard2019searching} demonstrate the effectiveness of multiple paths within a basic block. The DenseNet~\cite{huang2017densely} proposed by Huang et al. introduced dense connections between layers to improve feature reuse and alleviate the vanishing-gradient problems. Moreover, the Transformers-based networks, such as ViT~\cite{dosovitskiy2020image} in Fig.\ref{othernetwork} proposed by Dosovitskiy et al. and DETR~\cite{carion2020end} proposed by Carion et al., have achieved state-of-the-art performance on image classification and object detection tasks, respectively, by leveraging the self-attention mechanism. These novel visual networks have shown promising results and have opened up new research directions in the field of computer vision.

\section{Method}

\begin{figure}
    \centering
    \includegraphics[width=0.48\textwidth]{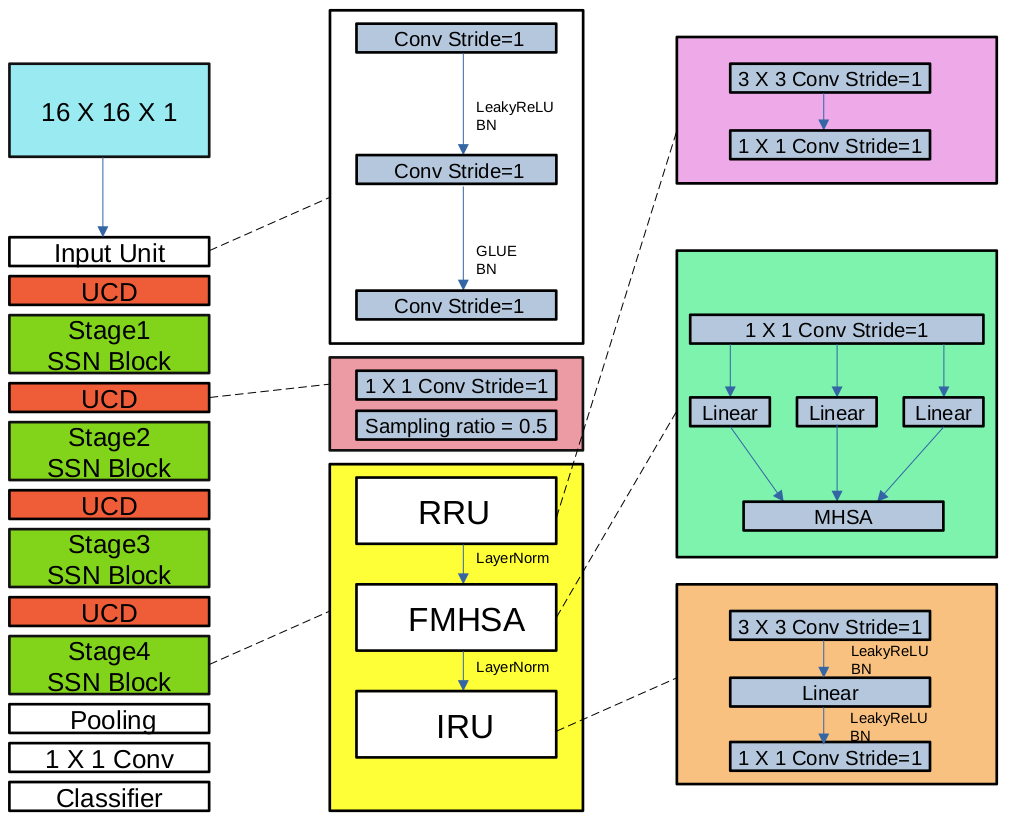}
    \caption{Structure of Sequential Spatial Network (SSN)}
    \label{fig:SSN}
\end{figure}

\subsection{Trajectory Prediction}
Trajectory prediction is an essential task in autonomous driving for collision avoidance. In this task, the goal is to predict the future trajectory of the vehicle based on its current and past states. This prediction allows the autonomous vehicle to plan its future path and avoid potential collisions with other objects in the environment. What we obtain from our model and dataset are X, Y axis position, and yaw. Since the frame slot is known, we can easily get the velocity that AV needs. Besides, after combing fiction between ground and wheels, wind resistance, and other physic parameters, we can calculate the acceleration that AV needs so that we can control the motor force just like human driving via an accelerograph. Moreover, yaw can provide a driving wheel adjustment as well. These processes compose the basic requirements for autonomous driving.

Some mathematical models such as kinematic or dynamic models can show these processes. The kinematic model assumes that the vehicle moves in a straight line with a constant velocity and acceleration. The equations of motion for the kinematic model can be represented as:
\begin{equation}
\begin{split}
x(t) &= x_0 + v_0 \cos(\theta_0) t + \frac{1}{2} a_x t^2, \\
y(t) &= y_0 + v_0 \sin(\theta_0) t + \frac{1}{2} a_y t^2, \\
\theta(t) &= \theta_0 + \omega t,
\end{split}
\end{equation}
where $x(t)$ and $y(t)$ are the position of the vehicle at time $t$ in the $x$ and $y$ axes, respectively, $v_0$ is the initial velocity, $\theta_0$ is the initial orientation, $a_x$ and $a_y$ are the acceleration in the $x$ and $y$ axes, respectively, and $\omega$ is the angular velocity.

On the other hand, the dynamic model takes into account the forces acting on the vehicle, such as friction, air resistance, and gravity. Wind resistance is an important factor to consider when predicting the trajectory of an autonomous vehicle. The force of wind resistance can be calculated using the following formula:
\begin{equation}
F_{wind} = \frac{1}{2}\rho v^2 C_d A,
\end{equation}
where $F_{wind}$ is the force of wind resistance, $\rho$ is the density of air, $v$ is the velocity of the vehicle, $C_d$ is the drag coefficient, and $A$ is the cross-sectional area of the vehicle. The drag coefficient and cross-sectional area can be experimentally determined for a specific vehicle.

To incorporate wind resistance into the trajectory prediction model, we can use the above formula to calculate the additional force that the vehicle must overcome. This can then be used to adjust the predicted trajectory accordingly.

The equations of motion for the total dynamic model can be represented as:
\begin{equation}
\begin{split}
m\frac{d^2 x}{dt^2} &= F_x - F_{friction} - F_{wind}, \\
m\frac{d^2 y}{dt^2} &= F_y - F_g - F_{friction} - F_{wind},
\end{split}
\end{equation}
where $m$ is the mass of the vehicle, $F_x$ and $F_y$ are the forces acting on the vehicle in the $x$ and $y$ axes, respectively, $F_{friction}$ and $F_{wind}$ are the forces due to friction and wind resistance, respectively, and $F_g$ is the force due to gravity.

To predict the future trajectory of an autonomous vehicle, we need to estimate the parameters of the kinematic or dynamic model based on the current and past states of the vehicle. This can be done using machine learning techniques such as regression or neural networks.

A popular neural network model for trajectory prediction in autonomous driving is the Long Short-Term Memory (LSTM) network. LSTM networks are a type of recurrent neural network that can capture long-term dependencies in sequential data. The LSTM network can take as input the current and past states of the vehicle and predict its future trajectory.

Given a sequence of input states ${s_1, s_2, ..., s_T}$, where $s_t$ represents the state of the vehicle at time $t$, an LSTM network can predict the future states $\hat{s}{t+1}, \hat{s}{t+2}, ..., \hat{s}_{t+k}$, where $k$ is the number of future time steps to predict.

The LSTM network consists of an input layer, a hidden layer, and an output layer. The input layer takes the sequence of input states as input, and the hidden layer contains recurrent connections that allow the network to maintain a memory of past states. The output layer produces the predicted future states. The LSTM network can be trained using backpropagation through time to minimize the prediction error.

Several studies have proposed different variations of the LSTM network for trajectory prediction in autonomous driving, such as the Social LSTM~\cite{alahi2016social}, the ConvLSTM~\cite{xingjian2015convolutional}, and the TrajGRU~\cite{Gupta_2018_ECCV}. These models have shown promising results in predicting the future trajectory of the vehicle and avoiding potential collisions with other objects in the environment.

Recent works have also explored the use of self-attention mechanisms to capture both local and global information for trajectory prediction in autonomous driving~\cite{li2021multi}. These models, such as the proposed SSN architecture, combine CNN and self-attention to capture spatiotemporal features from input data and improve the ability of sequentially related inputs.

In summary, trajectory prediction for collision avoidance in autonomous driving is a crucial task that can be addressed using recurrent neural networks such as LSTM or hybrid models that incorporate self-attention mechanisms to capture both local and global information.

\subsection{BIFR Block}
To address this sequential relation problem, we also propose a bi-directional frame-related (BIFR) block to build up relations between sequential frames, as shown in Fig.~\ref{fig:CSSN}. This block consists of one flatten layer, one bi-directional LSTM, and two fully connected layers. By utilizing bi-directional LSTMs, the information of both past and future frames can be incorporated into the current frame, which can effectively solve the problem of unreachability and safety of the predicted trajectory for AV.
\begin{figure}
    \centering
    \subfigure[former case]{
    \includegraphics[width=0.48\textwidth]{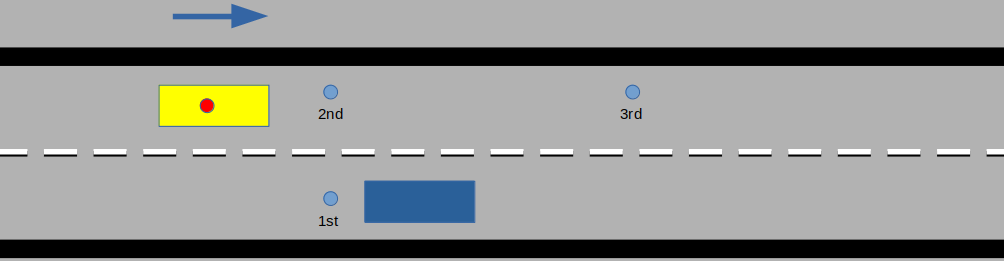}
    }
    \subfigure[later case]{
    \includegraphics[width=0.48\textwidth]{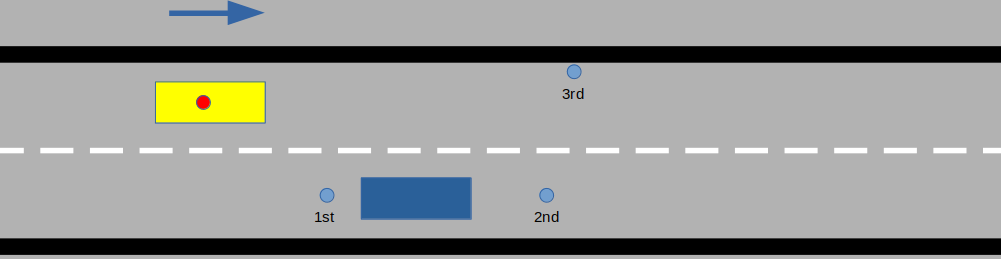}
    }
    \caption{Two different cases of predicted points can not reach}
    \label{fig:fomernot}
\end{figure}

The trajectory planning process is based on several waypoints, and these waypoints have a sequential relationship. The position of the later waypoints depends on the former waypoints, and the position of the former waypoints should consider the later waypoints as well. The potential danger of not considering the sequential relationship between waypoints can be observed from the examples shown in figures. The red circles represent the actual positions of the vehicle, and the blue circles represent the predicted positions based on the current waypoints. As can be seen from the figures shows in Fig.~\ref{fig:fomernot}, the former waypoints are reachable in the predicted trajectory but the latter waypoints are not, or vice versa. Therefore, it is essential to incorporate the sequential relationship between waypoints into the trajectory planning process to ensure the safety and feasibility of the trajectory.

\subsection{Network Structure Overview}

Our strategy is to take advantage of CNN, encoder, Bi-LSTM, and attention by building a hybrid network. CNN has excellent regional feature extraction capabilities, allowing the network to accurately and quickly capture important visual features within an image.
\begin{equation}
\mathbf{h}i = f(\sum{j=1}^{N} \mathbf{w}j * \mathbf{x}{i-j}  + b)
\mathbf{h}_i = f(\sum{j=1}^{N} \mathbf{w}_j * \mathbf{x}_{i-j} + b)
\end{equation}
Encoder can efficiently compress and encode the input data, reducing the computational complexity of the neural network.
\begin{equation}
Encoder = f_{enc}(x_1, x_2, ..., x_T).
\end{equation}

LSTM can learn and model long-term dependencies within sequential data, making it particularly effective in modeling and predicting temporal information.
\begin{equation}
\begin{aligned}
f_t &= \sigma_g(W_f x_t + U_f h_{t-1} + b_f), \\
i_t &= \sigma_g(W_i x_t + U_i h_{t-1} + b_i), \\
\tilde{C}t &= \sigma_c(W_c x_t + U_c h{t-1} + b_c), \\
C_t &= f_t \odot C_{t-1} + i_t \odot \tilde{C}t, \\
o_t &= \sigma_g(W_o x_t + U_o h{t-1} + b_o), \\
h_t &= o_t \odot \sigma_h(C_t).
\end{aligned}
\end{equation}

Attention mechanisms can dynamically weight different regions of an image or sequence, allowing the network to selectively focus on the most important information and improving overall performance.
\begin{equation}
\begin{aligned}
h_t = \sum_{i=1}^{T_x} \alpha_{t,i} \cdot {h}_i, \\
h_t = \sum_{i=1}^{T_x} \alpha _{t,i} \cdot {h}_i, \\
\alpha_{t,i} = \frac{\exp(e_{t,i})}{\sum_{j=1}^{T_x} \exp(e_{t,j})}, \\
e_{t,i} = f({s}_{t-1}, {h}_i).
\end{aligned}
\end{equation}

The combination of these techniques can allow the network to effectively model complex and dynamic relationships within a scene, allowing for more accurate and robust predictions.

In particular, the use of attention mechanisms can help the network to better handle variable and unpredictable inputs, allowing it to adapt to a wider range of scenarios and improve generalization performance. 

The bi-directional frame-related(BIFR) block is one of the fundamental building blocks in our networks, especially aims to capture sequential information in image data. The first step in the BIFR is to flatten the input sub-graphs, which converts the input data to a one-dimensional format that can be processed by the following layers. Next, the Bi-LSTM layer plays a central role in capturing local sequential information. By recording the temporal dependencies between the input data, the Bi-LSTM layer enables the network to learn the long-term dependencies between the input and output. To enhance the computational efficiency of the network, only the output value is kept, and the hidden layer parameters are ignored. This allows the subsequent layers to process the output value more efficiently. Finally, two fully connected layers are used to map the output value produced by the Bi-LSTM layer as inputs for later blocks, which further process the information to extract higher-level features. Overall, the BIFR is a powerful tool for capturing sequential information in image data and has the potential ability widely used in many state-of-the-art computer vision networks, especially for sequentially related inputs.

\subsection{SSN Blcok}
Resnet-50 is a powerful architecture that has proven to be effective for various image classification tasks. Its design comprises of five stages, with the initial stage0 consisting of a convolutional layer, a batch normalization layer, and a maxpooling layer. The bottleneck blocks used in stages 1-4 allow for efficient feature extraction and are followed by a fully connected layer for classification. One of the advantages of this design is that it enables efficient classification between different images. However, stage0 only uses one convolutional layer with a large kernel size, which can limit the capture of local information. To address this limitation, the main input block is introduced, which is composed of three different kernel sizes and steps convolutional layers. The main objective of the input block is to extract input information quickly when the output size is large and then use the smaller kernel sizes to capture local information as the input size decreases. This approach helps to improve the overall performance of our model.

The proposed SSN block consists of a Reinforcement Region Unit (RRU), a Fast Multi-Head Self-Attention (FMHSA) module and an Information Refinement Unit (IRU), as shown in Fig.~\ref{fig:SSN}. We will describe these four components in the following.
After taking into account the fast processing and local information processing of the main input block, the input information is transferred to the subsequent blocks for subsequent processing. Furthermore, between each block, we add a UCD layer, which consists of a convolutional layer with 1×1 kernel size and a downsampling layer which a sampling ratio is 0.5. The UCD layer allows us to speed up the network without reducing the amount of information in the input but maintaining the ratio between the information, and the size of the input is reduced to half of the original size after the UCD layer. Afterward, the feature extraction is performed by a network layer composed of different numbers of SSN blocks, while maintaining the same resolution of the input. Due to the existence of the self-attention mechanism, SSN can capture the correlation of different local information features, so as to achieve mutual dependence between local information and global information. Finally, the results are output through an average pooling layer and a projection layer as well as a classifier layer. 
The reinforcement region unit (RRU) was proposed as a data augmentation technique for our network. Data augmentation technique can generate high-quality augmented samples that maintain the spatial correlation and semantic information of the original image by selectively amplifying the informative regions of the input images. This technique can improve the robustness and generalization of the model, and it has been applied to various computer vision tasks such as object detection, semantic segmentation, and image classification. Moreover, RRU can be easily integrated with existing models and does not require additional computational resources, making it an efficient and practical data augmentation method for deep learning models.

In other words, a good model should maintain effective operating output for similar but variant data as well so that the model has better input acceptability. However, the absolute position encoding used in the common attention was originally designed to exploit the order of the tokens, but it breaks the input acceptability because each patch adds a unique position encoding to it ~\cite{chu2021conditional}. Furthermore, the concatenation between the local information obtained by the information capture module at the beginning of the model and the structural information inside the patch ~\cite{islam2020much} is ignored. In order to maintain input acceptability, the Reinforcement Region Unit (RRU) is designed to extract the local information from the input to the "SSN" module, defined as:
\begin{equation}
RRU(X)=Conv(Conv(X)).
\end{equation}

The FMHSA module in our model is designed to enhance the connection between local information obtained from the RRU. With the use of a convolutional layer, a linear layer, and multi-head self-attention, we can effectively capture the correlation between different local features. This is particularly useful for the collision avoidance task, where the trajectory prediction is based on continuously predicted positions that are sequentially related. The FMHSA module allows for the transfer of local information between different areas, making it suitable for solving this problem. By improving the connection between local information, our model is able to achieve outstanding results in this task. Additionally, the sequential relationship between predicted positions is taken into account, ensuring that the autonomous driving vehicle arrives at each target position before moving on to the next one.

The Information Refinement Unit (IRU) is used to efficiently extract the local information obtained by FMHSA, and after processing by this unit, the extracted local information is fed into the pooling and classifier layers. The original FFN proposed in ViT consists of two linear layers separated by the GELU activation~\cite{hendrycks2016gaussian}. First, expand the input dimension by 4 times, and then scale down the expanded dimension: 
\begin{equation}
FFN(X) = GELU(XW1 + b1)W2 + b2.
\end{equation}

This has the advantage of using a linear function for forward propagation before using the GELU() function, which greatly improves the efficiency of the model operation. The proposed design concept is to address the performance sacrifice issue of using the linear function for forward propagation in the FFN of ViT. The solution is to use a combination of convolutional layers with different kernel sizes and a linear function layer to achieve both operational efficiency and model performance. First, a larger convolutional kernel is used to capture the input information's characteristics with a large field of view. Then, the linear function layer is applied for fast propagation of the input information. Finally, a smaller convolutional kernel is used to refine the obtained information. By using this approach, the model can efficiently process the input information while maintaining high performance during fast propagation. This strategy is particularly useful for vision tasks such as image classification and object detection, where the model's performance and operational efficiency are crucial.
\begin{equation}
IRU(X)=Conv(L(Conv(X))),
\end{equation}
where L(X)=WX+b. After designing the above three unit modules, the SSN block can be formulated as: 
\begin{equation}
\begin{split}
    A&=RRU(X), \\ 
    B&=FMHSA(A), \\ 
    C&=IRU(B)+B.
\end{split}
\end{equation}

In the experiment part, we will prove the efficiency of SSN network.
In addition to its ability to capture the temporal dependencies of sequential data, SSN also has versatility in handling different downstream tasks. Specifically, SSN employs a self-attention mechanism to capture the relationships between the different temporal feature maps. This mechanism can effectively encode the dependencies between different timesteps and extract the most relevant information for the downstream tasks. Furthermore, SSN is designed to handle the sparsity and irregularity of the sequential data, which is common in real-world driving scenarios. These features make SSN a promising candidate for various applications in autonomous driving, such as trajectory prediction, behavior planning, and decision-making.


\section{Experiment}

The autonomous driving obstacle avoidance task is crucial in ensuring safe driving in real-life scenarios. To evaluate the effectiveness of the BCSSN architecture, experiments were conducted using a driving map as the main input. We use the Earlystopping function with 100 patience for finding the best parameters' value. Besides, we apply cosine dynamic learning rate~\cite{smith2018disciplined} adjustment for tuning suitable values at each iteration. To implement the cosine dynamic learning rate adjustment, we use the following formula:
\begin{equation}
\eta_t=\eta_{max}\frac{1+\cos(\frac{t\pi}{T})}{2}
\end{equation}
where $\eta_{max}$ is the maximum learning rate, $t$ is the current iteration, and $T$ is the total number of iterations.

The EarlyStopping function is used to monitor the validation loss and stop the training process when the validation loss does not improve for a certain number of epochs. Specifically, we use the patience parameter, which determines how many epochs to wait for the validation loss to improve before stopping the training process. The EarlyStopping function is defined as follows:
\begin{equation}
\text{EarlyStopping}(V_{\text{loss}},\text{patience})
\end{equation}
where $V_{\text{loss}}$ is the validation loss and $\text{patience}$ is the number of epochs to wait for the validation loss to improve.
 In the experiments, the proposed BCSSN architecture was compared with other popular models. The experimental results in Table~\ref{tab1} were then analyzed to draw conclusions on the performance of the BCSSN architecture. To assess the performance of the BCSSN architecture, collisions were defined into three categories: front collision, rear collision, and side collision. These collisions were caused by different unsuitable physical parameters, and they are depicted in Fig.~\ref{fig:collision}. The experiments aimed to test the ability of the BCSSN architecture to detect and avoid these collisions effectively. The results of the experiments were analyzed, and the performance of the BCSSN architecture was compared with other popular models to determine its effectiveness in addressing the obstacle avoidance task. The analytical conclusions drawn from these experiments provide valuable insights into the efficiency and efficacy of the proposed BCSSN architecture.

\subsection{Dataset and Description}

The l5kit dataset is a vast collection of data, which serves as the primary source for this study. It is an extensive dataset containing over 1,000 hours of data collected over a four-month period by a fleet of 20 autonomous vehicles that followed a fixed route in Palo Alto, California. The dataset comprises 170,000 scenes, where each scene lasts for 25 seconds. Each scene captures the perception output of the self-driving system, which provides precise positions and motions of nearby vehicles, cyclists, and pedestrians over time. This information is invaluable for training and testing autonomous driving models.

In addition to the scene data, the l5kit dataset also contains a high-definition semantic map with 15,242 labeled elements, which includes information about the lane markings, traffic signs, and other relevant information. This semantic map is used to help the autonomous vehicle understand the environment and make informed decisions. Furthermore, the dataset also includes a high-definition aerial view of the area, which provides a top-down perspective of the environment.

Overall, the l5kit dataset is an invaluable resource for researchers and developers working in the field of autonomous driving. With its vast collection of data, including the perception output of the self-driving system, semantic map, and aerial view, the dataset provides a comprehensive understanding of the driving environment, which can be used to train and test autonomous driving models.

\subsection{Data Preprocessing}

The l5kit dataset has a well-defined structure, consisting of three main concepts: Scenes, Frames, and Agents. Scenes are identified by the host vehicle that collected them and a start and end time. Each scene consists of multiple frames, which are snapshots captured at discretized time intervals. The frames are stored in an array, and the scene datatype stores a reference to its corresponding frames in terms of the start and end index within the frames array. All the frames in between these indices correspond to the scene, including the start index, but excluding the end index.

Each frame captures all the information observed at a particular time, including the timestamp, which the frame describes. The frame also includes data about the ego vehicle itself, such as its position and rotation. Additionally, the frame contains a reference to other agents (such as vehicles, cyclists, and pedestrians) that were detected by the ego's sensors. The frame also includes a reference to all traffic light faces for all visible lanes.

An agent in the l5kit dataset is an observation made by the autonomous vehicle of another detected object. Each agent entry describes the object in terms of its attributes, such as position and velocity, and gives the agent a tracking number to track it over multiple frames (but only within the same scene!) and its most probable label.

The input of this dataset is images of the Ego car, which is one of the properties of the EgoDataset. The output of the model is the position and yaw, which are also properties of the EgoDataset. By using this method, it is possible to simulate vehicles' driving as human driving actions. During the human driving process, drivers control the accelerator and the driving wheels to move the vehicles, where the accelerator is used for velocity and the driving wheel for yaw. Similarly, the output of the model is also the velocity and yaw, which can be used to simulate the trajectories of vehicles and avoid collisions during driving.

\subsection{Result}

We conducted experiments using four different network structures and presented their results in Table~\ref{tab1}. We compared our model's performance with other transformer-based and convolution-based models and found that our model achieved better accuracy and faster processing speed.

The results highlighted the significance of our proposed BCSSN for capturing both local and global information, which resulted in a significant reduction of front collision errors. Specifically, our model achieved 2.6 times fewer front collision errors compared to RepVGG, which is 13.6 times less than RepVGG's error rate. Similarly, our model achieved 5.8 times fewer front collision errors than ViT and 12.6 times less than ResNet50. Other comparisons can observe form result table~\ref{tab1} directly as well. These results demonstrate the effectiveness of the BCSSN in capturing both local and global information, leading to significant improvements in the model's accuracy.

BCSSN consistently outperformed other models by a large margin across all the test scenarios. These findings indicate that the proposed BCSSN block is a promising technique for improving the performance of autonomous driving systems by effectively capturing both local and global information.
\begin{table}[ht]
    \caption{Average collision times per 1000 miles.}
    \renewcommand\arraystretch{1.5}
    \centering
    \begin{tabular}{@{\hspace{0pt}}m{3cm}<{\centering}@{\hspace{0pt}}@{\hspace{0pt}}m{1.3cm}<{\centering}@{\hspace{0pt}}@{\hspace{0pt}}m{1.3cm}<{\centering}@{\hspace{0pt}}@{\hspace{0pt}}m{1.3cm}
    <{\centering}@{\hspace{0pt}}@{\hspace{0pt}}m{1.3cm}
    <{\centering}@{\hspace{0pt}}}
    \hline
    \hline
        Method & Front & Side & Rear & Total \\ 
         \hline
         BC~\cite{urban} &79 &395 &997 &1471 \\
         BC-perturb~\cite{urban} &14 &73 &678 &765 \\
         MS Prediction~\cite{urban} &18 &55 &125 &198 \\
         Urban Driver~\cite{urban} &15 &46 &101 &162 \\
         ML Planner~\cite{vitelli2022safetynet} & - & - & - &91.5\\
         L5Kit (ResNet-50)~\cite{houston2021one} &15.2 &20.7 &8.3 &44.2 \\
         RepVGG~\cite{ding2021repvgg} &16.2 &11.7 &10.6 &38.5 \\
         Vit~\cite{dosovitskiy2020image} &8.4 &\textbf{7.7} &9.2 &25.3 \\
         SafetyNet~\cite{vitelli2022safetynet} & - & - & - &4.6 \\
         BCSSN (Ours) &\textbf{0.6} &1.1 &\textbf{2.0} &3.7 \\
    \hline
    \hline
    \end{tabular}
    \label{tab1}
\end{table}

\section{Feature Work}
\begin{figure}
    \centering
    \subfigure[Bird's eye view of virtual city]{
    \includegraphics[width=0.48\textwidth]{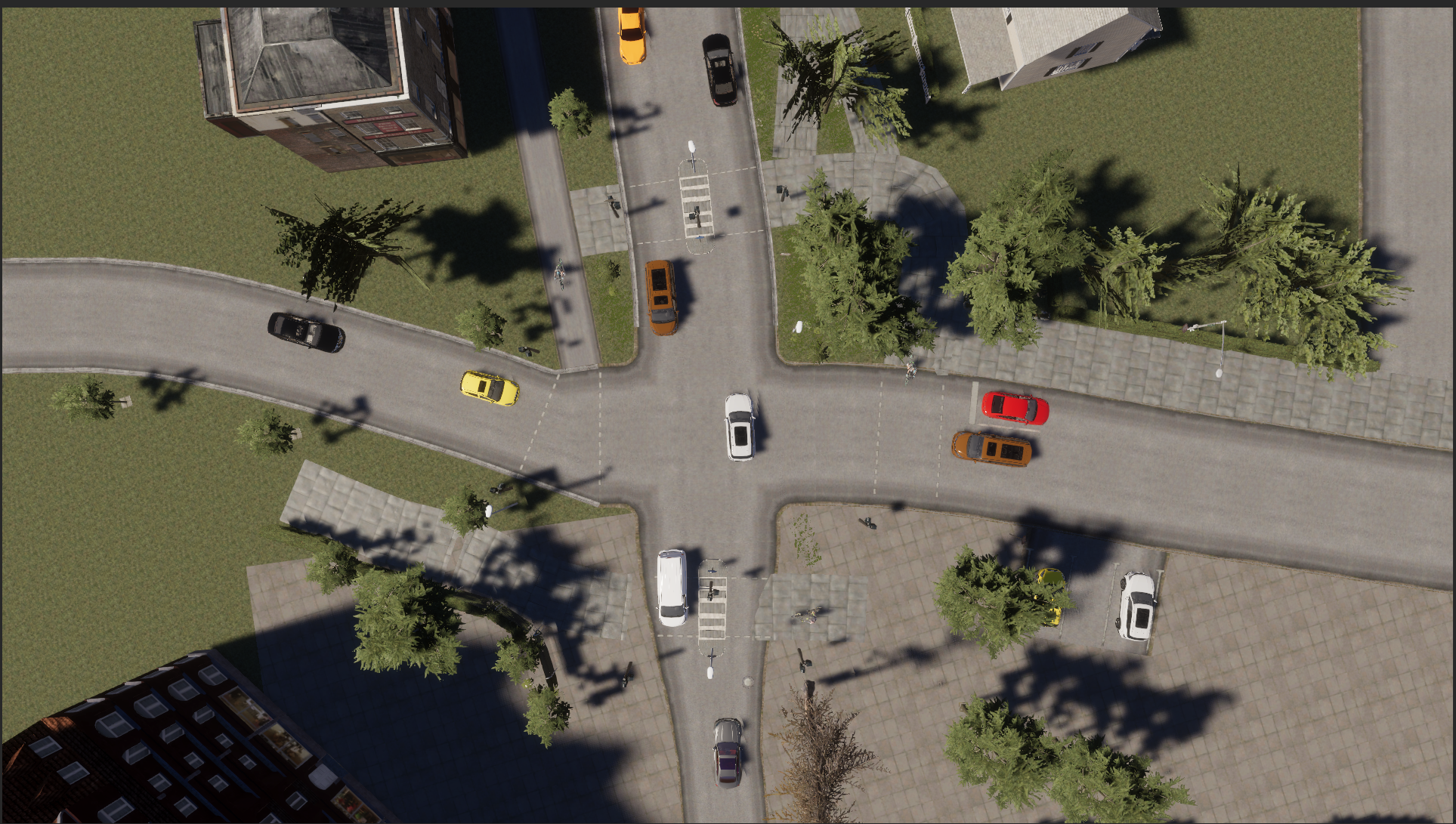}
    }
    \subfigure[First person view of virtual city]{
    \includegraphics[width=0.48\textwidth]{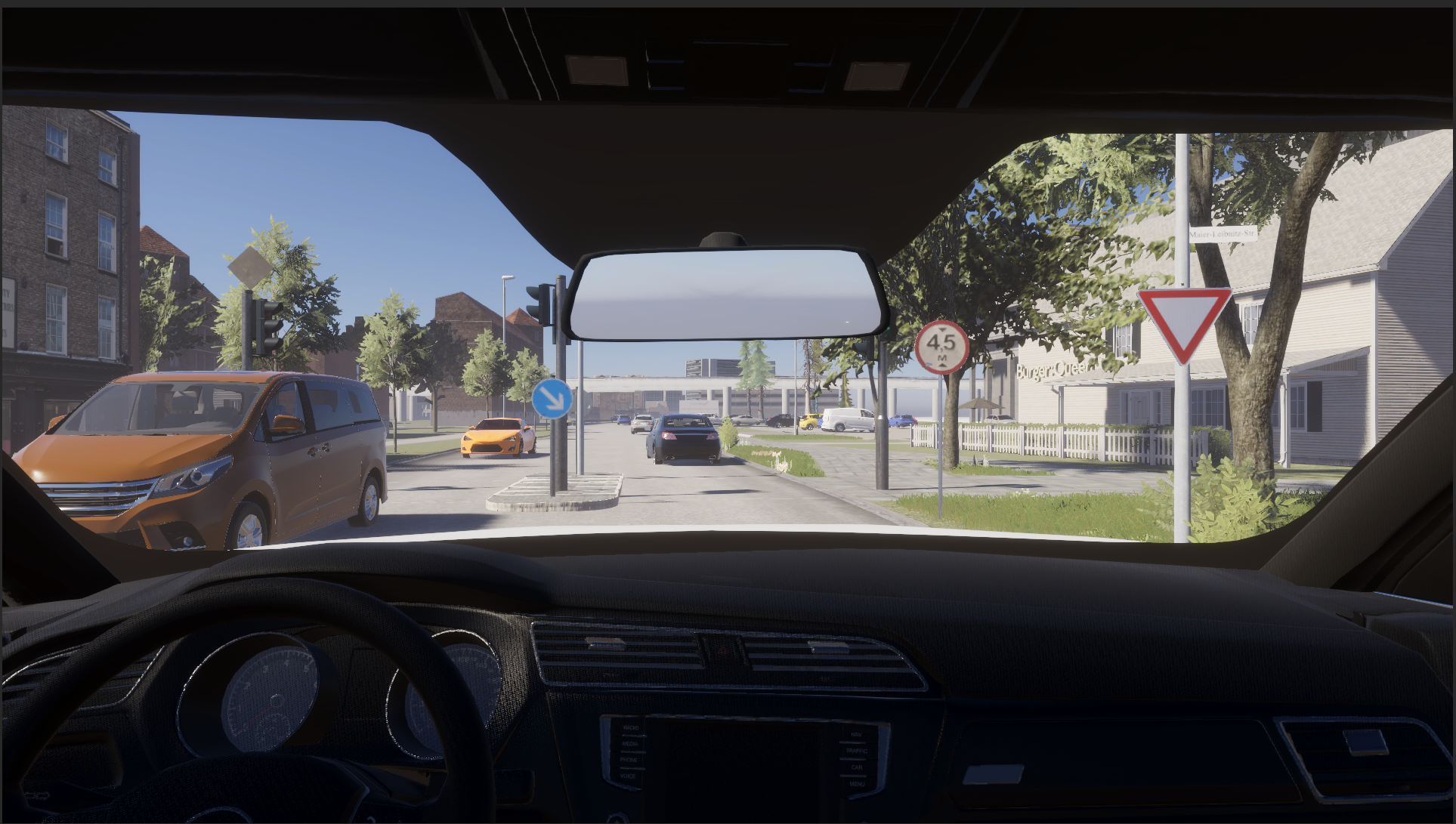}
    }
    \caption{Autonomous driving vehicle in Unity virtual city}
    \label{fig:first}
\end{figure}

To integrate our algorithm model with the Unity virtual world, we need to establish communication between them. We accomplish this by using Unity's built-in API and message-passing techniques, a first-person view, and bird eye view figure shown in Fig.~\ref{fig:first} . The communication protocol is established such that the algorithm sends the input information to the virtual world, and the virtual world returns the output information to the algorithm.

To ensure accurate physical modeling and simulation, we need to perform coordinate transformation between the image coordinates and the world coordinates. The process involves mapping the image coordinates to the virtual world coordinates using the camera parameters and the intrinsic and extrinsic calibration matrices. The resulting coordinates are then transformed to the physical world coordinates, where the physical model can calculate the necessary forces and physical parameters for the autonomous driving system.

This transformation can be mathematically represented as:
\begin{equation}
\begin{bmatrix}
X_{world}\\
Y_{world}\\
Z_{world}
\end{bmatrix} = R\begin{bmatrix}
X_{camera}\\
Y_{camera}\\
Z_{camera}
\end{bmatrix} 
+ T
\end{equation}
where $\mathbf{R}$ is the rotation matrix, $\mathbf{T}$ is the translation vector, and $[X_{camera}, Y_{camera}, Z_{camera}]$ are the coordinates in the camera frame.

To obtain accurate values for $\mathbf{R}$ and $\mathbf{T}$, we need to perform calibration using various methods such as checkerboard pattern or laser scanning techniques ~\cite{zhang2018evaluation}. The resulting calibration parameters can then be used to transform the image coordinates to world coordinates for accurate physical modeling and simulation.

\section{Conclusion}
In this paper, we introduce a novel hybrid architecture named BCSSN that is designed to improve the ability of vision-based autonomous driving tasks as well as other vision tasks. The BCSSN architecture combines the strengths of several different techniques, including CNN, Bi-LSTM, encoders, and self-attention, to capture both local and global information from sequentially related inputs.

The CNN are used to extract local information from the input images, while the Bi-LSTM is used to model the temporal dependency between the frames. The encoder is used to convert the input images into a high-level feature representation, which is then passed to the self-attention mechanism to capture the long-range dependencies and context information.

To evaluate the effectiveness of the proposed BCSSN architecture, we conducted extensive experiments on the Lykit dataset, which is a challenging dataset for vision-based autonomous driving tasks. The results of these experiments demonstrate the superiority of the proposed BCSSN architecture compared to other state-of-the-art models. Specifically, the BCSSN architecture achieved better performance in terms of accuracy, speed, and generalization ability.

Overall, BCSSN architecture offers a promising solution for vision-based autonomous driving tasks and other vision tasks that require the ability to capture both local and global information from sequentially related inputs. Its hybrid design allows it to take advantage of the strengths of several different techniques, resulting in improved performance and generalization ability.


{\small
\bibliographystyle{unsrt}

\bibliography{ref}
}

\end{document}